\documentclass[conference]{IEEEtran}
\IEEEoverridecommandlockouts
\usepackage{cite}
\usepackage{amsmath,amssymb,amsfonts}
\usepackage{algorithmic}
\usepackage{graphicx}
\usepackage{textcomp}
\usepackage{xcolor}
\usepackage{booktabs}  % For \toprule, \midrule 和 \bottomrule
\usepackage{color}     % For \color命令
\usepackage{colortbl}  % For \rowcolor命令
\usepackage{hyperref}
\usepackage{subcaption}
\usepackage{graphicx}
\usepackage{hyperref}
\hypersetup{
    colorlinks=true,
    linkcolor=red,
    filecolor=green,      
    urlcolor=blue,
    citecolor=green,
}

\def\BibTeX{{\rm B\kern-.05em{\sc i\kern-.025em b}\kern-.08em
    T\kern-.1667em\lower.7ex\hbox{E}\kern-.125emX}}
\begin{document}

\title{
Authentic Emotion Mapping: Benchmarking Facial Expressions in Real News
}

\author{Qixuan Zhang$^{1}$\thanks{Email: \textit{\{qixuan.zhang, zhifeng.wang, kaihao.zhang, sabrina.caldwell\}@anu.edu.au}}, Zhifeng Wang$^{1}$\thanks{\textit{\{lyf1082, kf.zy.qin\}@gmail.com}, \textit{tom.gedeon@curtin.edu.au}}, Yang Liu$^{1}$, Zhenyue Qin$^{1}$, Kaihao Zhang$^{1}$,  Sabrina Caldwell$^{1}$, Tom Gedeon$^{2}$\\
Australian National University$^1$, Curtin University$^2$\\
%{\tt\small \{yang.liu3, zhenyue.qin\}@anu.edu.au} \\
% \tt\small saeed.anwar@csiro.au, peterji530@gmail.com, dongwookim@postech.ac.kr}
}
\maketitle

\begin{abstract}
In this paper, we present a novel benchmark for Emotion Recognition using facial landmarks extracted from realistic news videos. Traditional methods relying on RGB images are resource-intensive, whereas our approach with Facial Landmark Emotion Recognition (FLER) offers a simplified yet effective alternative. By leveraging Graph Neural Networks (GNNs) to analyze the geometric and spatial relationships of facial landmarks, our method enhances the understanding and accuracy of emotion recognition. We discuss the advancements and challenges in deep learning techniques for emotion recognition, particularly focusing on Graph Neural Networks (GNNs) and Transformers. Our experimental results demonstrate the viability and potential of our dataset as a benchmark, setting a new direction for future research in emotion recognition technologies. The codes and models are at:
 \url{https://github.com/wangzhifengharrison/benchmark_real_news}

\end{abstract}


\section{Introduction}
Emotion recognition has become a part of interactive technology, finding its place in numerous applications ranging from content delivery \cite{saxena2020emotion} to medical diagnostics \cite{chen2020interpretation} and immersive virtual experiences \cite{marin2020emotion}. Traditional approaches to emotion recognition have heavily relied on RGB images, demanding significant processing resources to analyze complex facial features in such high-resolution data \cite{ngoc2020facial}, \cite{zhang2017facial}.

Facial Landmark Emotion Recognition (FLER) presents an alternative that simplifies this process through the use of basic geometric constructs on the face \cite{chen2021residual}.

\subsection{Introduction to Facial Landmark Emotion Recognition}

Emotion recognition through facial expressions is a significant aspect of artificial intelligence, aiming to interpret the emotional states by human facial gestures. The use of facial landmarks, or distinct facial points, is essential in this domain, enabling the detection of subtle emotional variations. The subtle shifts in these landmarks are closely linked to different emotional states and their precise recognition is critical for enhancing the performance of emotion recognition systems. Compared to traditional image-based emotion recognition, facial landmark-based emotion recognition can protect privacy since it does not involve processing entire facial images \cite{chen2021residual}. Moreover, in resource-constrained scenarios like edge computing, landmark-based methods are favored for their efficiency and low resource consumption. Consequently, Facial Landmark Emotion Recognition (FLER) has emerged as a pivotal technology across various applications, prominently in smart devices and real-time interactive systems \cite{poulose2021feature}.

\subsection{Graph Neural Networks in Emotion Recognition}

The Graph Neural Networks (GNNs) \cite{jin2021learning}, \cite{zhong2019graph} has provided novel methods to handle complex structured graph data. In the domain of facial landmark emotion recognition, the GNNs are especially valuable because they can directly work with graphs composed of facial landmarks, where the edges define the structure of facial expressions. GNNs are adept at capturing the intricate interactions among facial landmarks, which are essential for understanding the facial dynamics associated with different emotional states. Although the application of GNNs in facial emotion recognition is still in its nascent stages, their theoretical advantages in handling graph-structured data suggest immense potential. For example, \cite{ngoc2020facial} introduces a directed graph neural network (DGNN) using graph convolution and landmark features for Facial Emotion Recognition (FER), efficiently capturing geometric and temporal information from facial landmarks and addressing the vanishing gradient issue through a stable temporal block. Rao \textit{et al.} \cite{rao2021facial} presents a novel multiscale graph convolutional network based on facial landmark graphs for Facial Expression Recognition (FER), addressing issues of data redundancy and bias. However, the deployment of GNNs in this field also presents challenges such as how to design effective graph structures to accurately reflect the dynamics of facial expressions, and balancing the complexity of the models with the performance constraints of edge computing devices \cite{zhao2021geometry}.

\subsection{Gaps in Current Research and Our Contribution}

Despite numerous studies in the field of facial landmark emotion recognition, there is still a lack of a systematic benchmarking framework for evaluating and comparing different neural network models. Such a benchmark is vital for understanding how various models perform in real-world applications. Most existing studies have focused on specific models or algorithms without providing a comprehensive perspective on the performance of different methods when dealing with facial landmark data. Furthermore, the absence of benchmarking makes it difficult for researchers and developers to determine which models are best suited for deployment on resource-limited edge devices. To address these issues, this paper presents a benchmarking test, including a range of neural network models and their performance evaluations across different emotion classification tasks. We consider not only the accuracy of the models but also their computational efficiency, thus providing significant references for applications on edge computing devices. The motivation behind this study is to advance model performance and facilitate practical applications on edge devices. We believe that through these benchmarks, the research community can gain a deeper understanding of the strengths and limitations of different algorithms, leading to the design of solutions that are more in tune with practical requirements. 

The contribution of this paper lies in establishing a comprehensive and systematic benchmarking for facial landmark emotion recognition. We ensure the reliability and consistency of our test results through strict experimental design, providing a solid benchmark for assessing the performance of various neural network models in the task of facial emotion recognition. Furthermore, we delve into how the number and configuration of landmarks impact model performance, revealing the significance of landmark selection in optimizing model efficiency. Additionally, we have created our dataset and benchmarking code open source for the community to facilitate further research and development. These contributions not only provide the academic community with research resources but also lay a solid foundation for the industry to implement efficient algorithms, thereby promoting the development of technology geared towards practical applications.
\section{Related Work}
In this section, we explore the landscape of emotion recognition, starting from its fundamental tasks and the role of facial landmarks, to the latest advancements in deep learning, particularly focusing on Convolutional Neural Networks (CNNs) and Transformers. Each subsection delves into significant developments, challenges, and potential future directions, laying out a comprehensive overview of the field's progression and the cutting-edge techniques shaping the future of facial landmark-based emotion recognition.
\subsection{Introduction to Emotion Recognition and Facial Landmarks}
The journey toward automating emotion recognition has covered various fields, from early psychophysiological studies to the current artificial intelligence techniques. This section will introduce significant past works, specifically focusing on the use of facial landmarks as indicators for emotion recognition. Traditional emotion recognition systems mainly utilized methods based on geometric features, employing distances and angles between facial landmarks as direct inputs to classifiers such as Support Vector Machines (SVMs) \cite{liu2020ga} and simple neural networks\cite{hasani2017facial}.

\subsection{Advancements in Deep Learning for Emotion Recognition}
With the revival of neural networks, especially Convolutional Neural Networks (CNNs), the focus shifted to deeper neural networks for emotion recognition. Yang \textit{et al.} \cite{yang2017cnn} introduces a 3D facial expression recognition algorithm using CNNs and landmark features, robust to pose and lighting changes due to its reliance on 3D geometric models. Akhand \textit{et al.} \cite{akhand2021facial} introduces a highly accurate facial emotion recognition system employing a deep CNN model optimized via transfer learning and a novel pipeline strategy, addressing limitations of shallow networks and frontal-only images by fine-tuning with emotion data for enhanced feature extraction. However, these techniques often overlooked facial expressions' temporal dynamics. Addressing this, Kollias \textit{et al.} \cite{kollias2020exploiting} introduces a CNN-RNN approach for dimensional emotion recognition using multiple features and multi-task frameworks on large emotion datasets, significantly outperforming existing methods, but these still fell short in capturing the long-range dependencies crucial for understanding emotional expressions.

\subsection{Emotion Recognition with Transformers}
Enter the era of Transformers, which have redefined the possibilities in handling sequential data thanks to their self-attention mechanisms. Vaswani \textit{et al.} \cite{vaswani2017attention} revolutionized sequence modeling with the introduction of the Transformer model, which has since been adapted beyond the boundaries of language processing. In the realm of facial emotion recognition, recent studies have begun to explore the application of Transformers. For instance, Zhao \textit{et al.} \cite{zhao2022spatial} introduces a geometry-guided FER framework using graph convolutional networks and transformers, enhancing emotion recognition from videos by constructing spatial-temporal graphs with facial landmarks and incorporating attention mechanisms for more informative feature emphasis. Zheng \textit{et al.} \cite{zheng2023poster} introduces POSTER, a two-stream Pyramid cross-fusion Transformer network, aiming to address the key FER challenges: inter-class similarity, intra-class discrepancy, and scale sensitivity by fusing facial landmark and image features and employing a pyramid structure for scale invariance. Hybrid models have also emerged, combining the strengths of CNNs and Transformers. Karatay \textit{et al.} \cite{karatay2023cnn} presents a deep neural network framework combining Gaussian mixture models with CNN and Transformer for emotion detection from videos and images using facial and body features extracted by OpenPose, addressing various basic emotions.

\subsection{Addressing Diversity and Data Augmentation in Emotion Recognition}
Moreover, the impact of data diversity and representativeness has also been a focal point in the literature. Studies like that of Li \textit{et al.} \cite{li2020deep} highlighted the challenges posed by variations in ethnicity, age, and lighting conditions, prompting the development of more robust models. The employment of Generative Adversarial Networks (GANs) for data augmentation, as explored by Hajarolasvadi \textit{et al.} \cite{hajarolasvadi2020generative}, has been one avenue to enhance the diversity and quantity of training data, thereby improving the generalizability of emotion recognition systems.

\subsection{Conclusion and Future Directions in Facial Landmark-based Emotion Recognition}
The review of related work underscores a gradual but significant shift towards sophisticated models that consider both the spatial and temporal aspects of facial landmarks in emotion recognition. The exploration into Transformer-based models marks a new frontier in this domain, promising enhanced accuracy and deeper understanding of emotional states through the advanced modeling of sequential landmark data. This paper builds upon these foundations, aiming to further the efficacy and applicability of facial landmark-based emotion recognition systems.
%%%%% table 
\begin{table}[t]
\centering
\caption{Emotion Distribution Summary.}
\label{emotion_distribution_summary}
\renewcommand{\arraystretch}{1}
\setlength{\tabcolsep}{6pt} % Default value: 6pt
\begin{tabular}{lcccc}
\toprule
\textbf{Emotion} & \textbf{Training} & \textbf{Validation} & \textbf{Test} & \textbf{Total} \\
\midrule
\small
Angry & 914 & 100 & 200 & 1,214 \\
Fear & 2,031 & 100 & 200 &2,331\\
Happy & 3,091 & 100 & 200 & 3,391 \\
Neutral & 3,536 & 100 & 200 & 3,836 \\
Sad & 3,100 & 100 & 200 & 3,400 \\
Total & 12,672 & 500 & 1,000 & 14,172 \\
\bottomrule
\end{tabular}%
\end{table}
\section{Facial expressions in real news dataset}
\subsection{Dataset overview}
In Table \ref{emotion_distribution_summary}, striking feature of this distribution is the significant variance in the number of samples across different emotions. For instance, Neutral, Happy, and Fear emotions are notably more represented in the dataset, with counts exceeding 3,500, 3,000, and 2,000 respectively in the Training subset. The Total column sums the Training and Validation and Test samples, offering a holistic view of the dataset's composition, with a grand total of 14,172 instances. This comprehensive distribution is critical for understanding the dataset's balance and evaluating the potential biases in the training, validating testing of emotion recognition models.

\subsection{Basic emotion categories}
\begin{figure}
    \centering
    \includegraphics[width=0.9\linewidth]{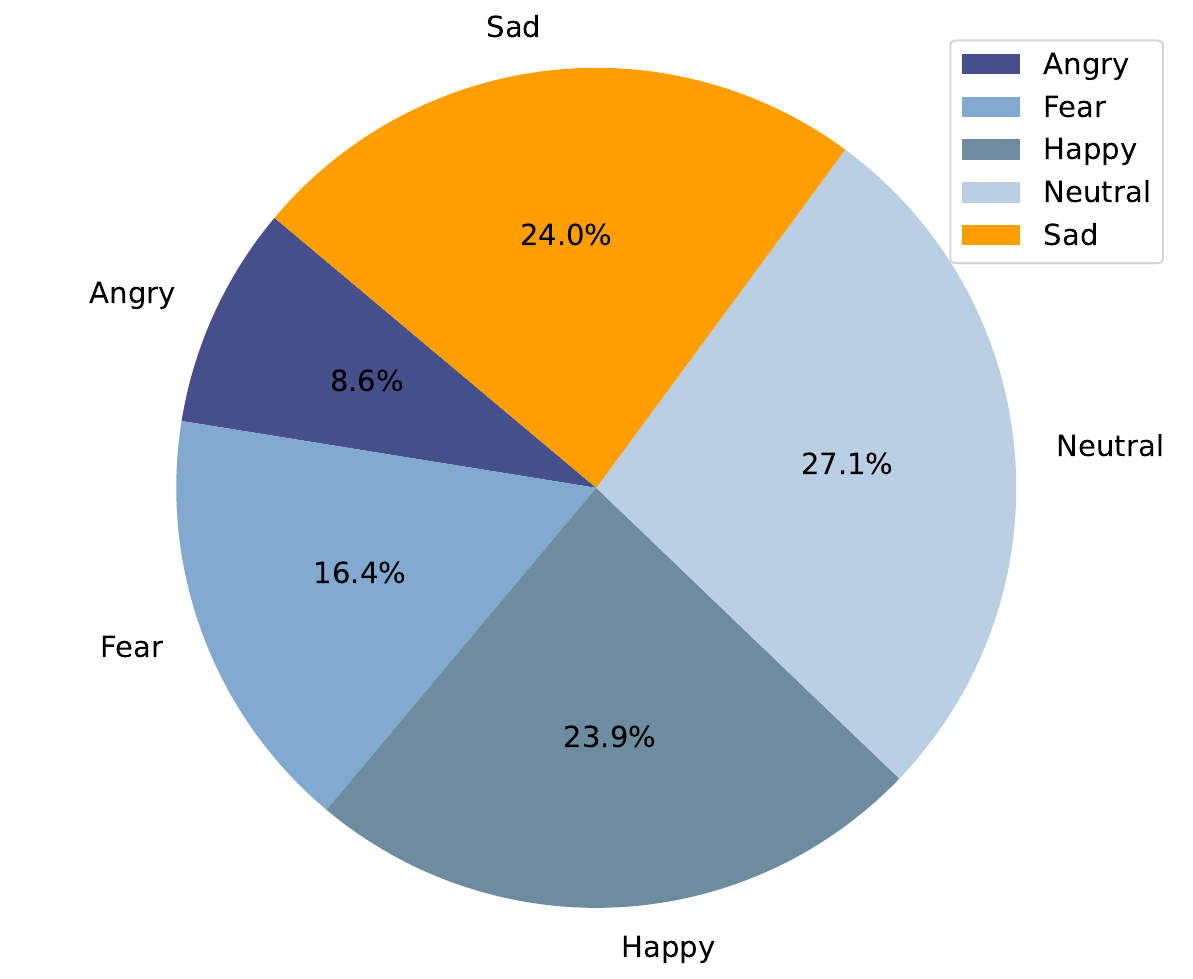}
    \caption{The number of emotion categories.}
    \label{pie_chart_for_emotion_categories}
\end{figure}

We capture 5 different daily basic emotion categories in real news. Fig.\ref{pie_chart_for_emotion_categories} illustrates the distribution of 5 different emotion categories, each represented by a unique color. The largest portion is `Neutral' at 27.1\%, followed closely by `Sad' at 24.0\%. `Happy' represents 23.9\% of the chart, while `Fear' is at 16.4\%. The smallest slice is `Angry', making up 8.6\% of the chart. Each category's percentage is provided.

%%%%% facial landmarks
\begin{figure*}
    \centering
    \includegraphics[width=0.9\linewidth]{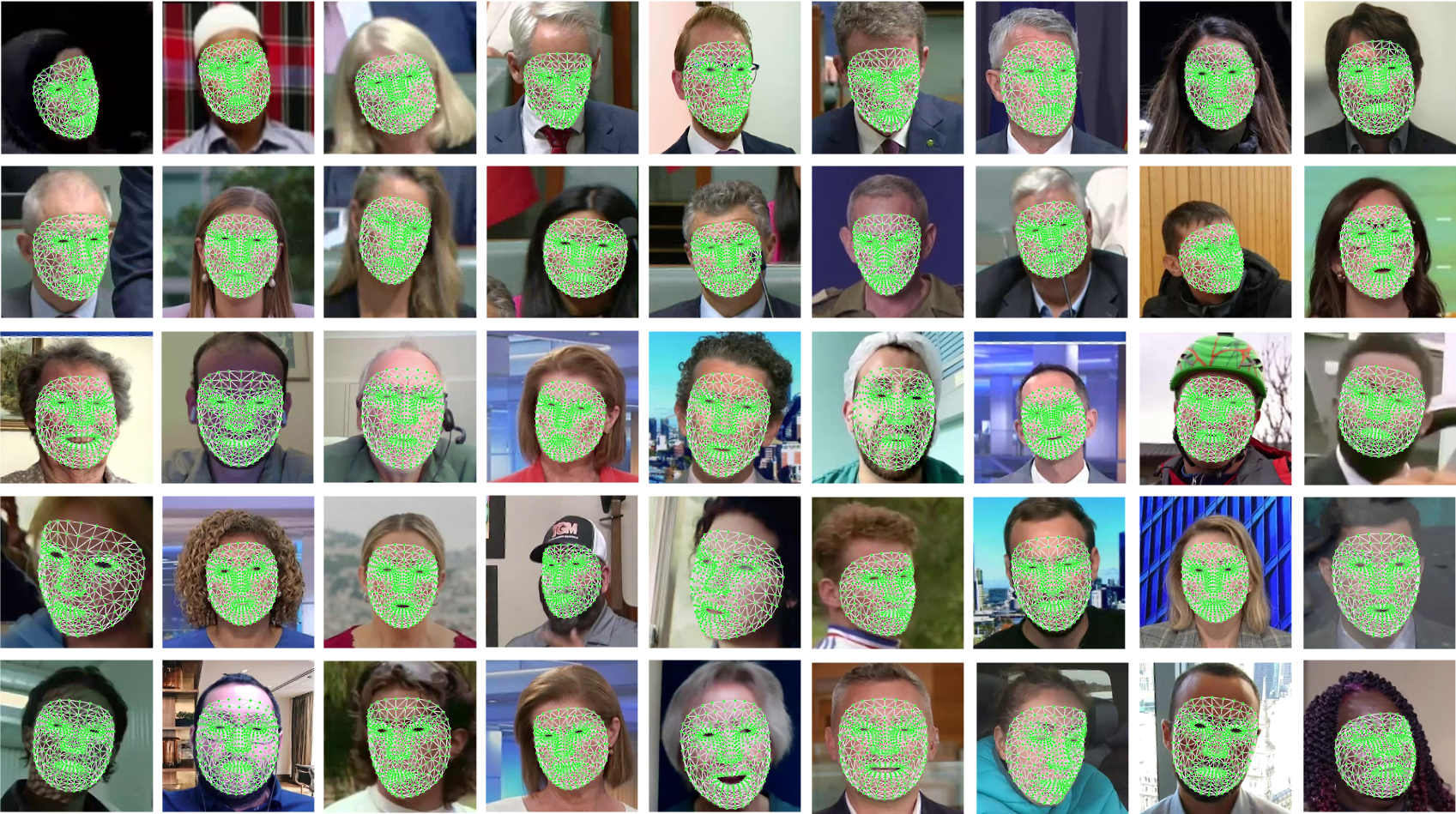}
    \caption{Landmarks faces}
    \label{transformer_attention}
\end{figure*}

% \begin{figure}
%     \centering
%     \includegraphics[width=0.9\linewidth]{2023wowimgs/stack_chart_v2.png}
%     \caption{Train and test distribution by category  }
%     \label{transformer_attention}
% \end{figure}

\subsection{Data acquisition and preparation}
We initiated our dataset by downloading 318 news videos from internationally recognized media sources such as ABC, Al Jazeera, NBC News, BBC, CBC, and CCTV and involved extracting clear, well-defined facial images from these videos. We established a quality control process that included evaluating image sharpness, ensuring all facial features were visible and distinct, and preventing the selection of repetitive images. Additionally, we employed adaptive padding technology to adjust each image's frame size based on the facial dimensions, thereby enhancing the overall quality and consistency of our image dataset.

To efficiently process and extract faces, we developed scripts to organize data into a structured directory, calculate adaptive padding, and use the Laplacian method for sharpness and mean squared error for image similarity. We also used Dlib's\cite{king2009dlib} frontal face detector to identify each face and extract its key features, making sure all essential landmarks were fully captured and within the image frame. This method ensured efficient and high-quality face extraction.

Our comprehensive facial analysis tool is a product of these efforts, adeptly identifying faces, pinpointing facial features, and recognizing emotional expressions. We utilized libraries such as DeepFace \cite{taigman2014deepface} and Mediapipe \cite{lugaresi2019mediapipe} to detect faces, analyze facial characteristics, and determine prevalent emotional states. Each piece of analysis was precisely documented, offering a rich dataset for future machine learning applications or detailed statistical analysis. This tool marks a significant advancement in the field of facial analysis.

We manually reviewed the images and corresponding landmark data, removing any irrelevant or low-quality images. After ensuring the integrity and relevance of the data, we shuffled and divided it into training and test sets, maintaining a balanced representation of each emotion category. A 'type' column was added to distinguish between the sets, laying the groundwork for a robust and reliable facial landmark emotion recognition benchmark. This structured, detailed approach to data acquisition and processing ensures a high-quality, diverse dataset that will significantly contribute to advancements in emotion recognition and facial analysis.

\section{Methods}

% \begin{figure}
%     \centering
%     \includegraphics[width=0.9\linewidth]{2023wowimgs/table_chart_v2.png}
%     \caption{Six emotional landscapes on transformer attention. }
%     \label{transformer_attention}
% \end{figure}

% Facial landmark emotion recognition is a domain where the application of Graph Neural Networks (GNNs) can be particularly impactful due to the inherently structured nature of facial features. The interconnectivity and spatial relations between facial landmarks (such as the corners of the mouth, the edge of the eyebrows, or the tip of the nose) are graph-like by nature, making GNNs an ideal candidate for capturing the subtle configurations that convey different emotions.

In the field of emotion recognition via facial landmarks, leveraging Graph Neural Networks (GNNs) offers significant potential. This stems from the structured arrangement of facial features. Facial landmarks, including mouth corners, eyebrow edges, and nose tips, naturally form a graph-like network. Their interconnectedness and spatial relationships are key to understanding emotional expressions. GNNs excel in this context, adeptly identifying the nuanced arrangements of these landmarks that represent various emotions.

\subsection{Innovative Convolution Techniques in Graph Neural Networks}

% Graph Convolutional Networks (GCNs) and Chebyshev Spectral CNNs (ChebNets) exemplify the use of convolution-based techniques in graph neural networks to interpret the complex spatial relationships between facial landmarks. GCNs can capture the nuances of emotional expressions through localized feature learning, while ChebNets utilize spectral graph theory to extract features based on the frequency of landmark configurations. Together, these methods offer a comprehensive approach to understanding both the static and dynamic aspects of facial expressions by integrating spatial and frequency domain analysis.

The application of convolution techniques in graph neural networks, particularly through Graph Convolutional Networks (GCNs)\cite{nguyen2022facial} and Chebyshev Spectral CNNs (ChebNets)\cite{li2020shape}, has been pivotal in decoding complex spatial interactions among facial landmarks. GCNs contribute by learning localized features, capturing the subtleties of emotional expressions. In contrast, ChebNets employ spectral graph theory, focusing on landmark configuration frequencies for feature extraction. Collectively, these methods provide a thorough framework for analyzing facial expressions, integrating both spatial and frequency domain insights to interpret static and dynamic expression elements.

\subsection{Attention and Aggregation Strategies for Enhanced Feature Analysis}

% Graph Attention Networks (GATs) and Graph Sample and Aggregate (GraphSAGE) networks both focus on the significance of individual landmarks but with different strategies. GATs leverage attention mechanisms to prioritize critical features dynamically, enhancing the model's focus on the most expressive facial regions. GraphSAGE, on the other hand, introduces a robust aggregation framework that generalizes across different facial structures and expressions, ensuring consistent performance even with diverse facial data.

Graph Attention Networks (GATs)\cite{prados2022shape} and Graph Sample and Aggregate (GraphSAGE) \cite{hamilton2017inductive}  networks adopt distinct approaches to emphasize the importance of individual facial landmarks. GATs utilize attention mechanisms to dynamically highlight key features, thereby improving the model's ability to focus on facial regions most indicative of emotions. Conversely, GraphSAGE introduces a versatile aggregation framework. This framework is adaptable to various facial structures and expressions, ensuring stable performance across a wide range of facial data.

\subsection{Dynamic and Edge-Enhanced Learning for Expressive Feature Interpretation}

% Dynamic Graph CNNs (DGCNNs) and Edge-Conditioned Convolution (ECC) networks offer advanced methods for interpreting the changing nature of facial expressions. DGCNNs are tailored for temporal data, capturing the progression of emotional cues over time, which is essential in video-based emotion recognition. ECC networks augment this by considering the edge attributes between landmarks, thus enriching the model's understanding of how landmark relationships contribute to emotional expression.

Dynamic Graph CNNs (DGCNNs)\cite{liu2022graph} and Edge-Conditioned Convolution (ECC)\cite{han2022devil} networks present innovative approaches for analyzing the evolving characteristics of facial expressions. DGCNNs are specifically designed for time-series data, effectively tracking the development of emotional indicators over time, a critical aspect in video-based emotion recognition. ECC networks complement this approach by factoring in the attributes of edges connecting facial landmarks. This addition enhances the model's insight into the impact of landmark interrelations on the expression of emotions.

\subsection{Spatial-Temporal and Topological Insights with Advanced GNN Architectures}

% Spatial Graph Convolutional Networks and Graph Isomorphism Networks (GIN) provide insights into the facial landmark structure by leveraging spatial-temporal and topological data. Spatial networks excel in learning directly from the arrangement of landmarks, mapping out the spatial configurations that characterize different emotions. GIN, in contrast, discerns the underlying topological structures of these configurations, ensuring that the model can differentiate between similar but distinct emotional states by focusing on the arrangement and features of the landmark's neighbors.

Spatial Graph Convolutional Networks\cite{bai2021two} and Graph Isomorphism Networks (GIN)\cite{xu2018powerful} offer novel perspectives in understanding facial landmark structures, utilizing both spatial-temporal and topological data. Spatial networks are adept at directly learning from the distribution of landmarks, effectively identifying the spatial patterns associated with various emotions. Conversely, GIN focuses on the intrinsic topological structures of these patterns. This approach enables the model to distinguish between closely related yet distinct emotional states by analyzing the configuration and characteristics of neighboring landmarks.

\section{Experiments}

%%%%%% 5 emotion categories
% \subsection{Datasets}
\subsection{Experiment settings}
We executed experiments on our proposed dataset to evaluate 5 classes of expressions, utilizing an Ubuntu 20.04 workstation equipped with an NVIDIA GeForce GTX 3090Ti GPU. Network implementation was carried out using the Pytorch framework.
\subsection{Accuracy}
Accuracy is calculated by the formula:
\begin{equation}
\text{Accuracy (\%)} = \left( \frac{P}{N} \right) \times 100
\end{equation}
where \( P \) denotes the number of correct predictions, and \( N \) is the total observations for each emotion category.
\begin{table}[t]
\centering
\caption{Emotion recognition performance by different evaluated approaches on our proposed dataset.}
\label{accuracy_for_five_emotion}
\begin{tabular}{lcccc}
\toprule
\textbf{Emotion} & \textbf{MLP}\cite{palo2015use} & \textbf{GINFormer}\cite{rampasek2022GPS}& \textbf{GIN}\cite{xu2018powerful} &\textbf{SAGE}\cite{hamilton2017inductive}\\
\midrule
Ave Acc & 28.00\% & 33.10\% & 30.20\% & 32.80\%\\
\bottomrule
\end{tabular}
\end{table}

\subsection{Experimental Results}
Multilayer Perceptron (MLP)\cite{palo2015use} is a fundamental type of neural network architecture that has been extensively applied in the field of machine learning. As a form of a feedforward artificial neural network, MLP consists of an input layer, multiple hidden layers, and an output layer. In our experiments, we use one hidden layer as baseline for comparison with other methods. Each node, except for the input nodes, is a neuron with a RELU activation function. MLP utilizes backpropagation for training the network. In our proposed dataset, MLP achieved average score of 28.00\% on five basic emotion categories shown in Table \ref{accuracy_for_five_emotion}. Compared with GIN \cite{xu2018powerful}, MLP can't capture the intrinsic topological structures of facial landmarks. GIN enables the model to distinguish between closely related yet distinct emotional states by analyzing the configuration and characteristics of neighboring landmarks. So, GIN achieved higher accuracy with average score of 30.20\%. SAGE \cite{hamilton2017inductive} is a novel inductive framework for facial landmarks embedding that leverages facial landmark features to learn a function capable of generalizing to unseen facial landmarks, distinguishing it from matrix factorization-based methods and allowing simultaneous learning of neighborhood topology and facial landmarks distribution, applicable to both feature-rich and standard structural graphs. Compared with GIN, SAGE can learn the distribution of facial landmarks and give prediction of unseen facial landmarks which are important for unseen emotion categories in the inference process. So, SAGE achieve higher average accuracy with score of 
32.80\%. In order to achieve linear complexity,  GINFormer \cite{rampasek2022GPS} proposes a Graph Transformer, adept at handling diverse benchmarks. This paper provides a foundational understanding of positional and structural encodings, categorizing them into local, global, and relative types. They introduce a novel architecture that scales linearly with the graph size, serving as a universal function approximator. The proposed framework integrates three core elements—positional/structural encoding, local message-passing, and global attention—to achieve state-of-the-art results in graph representation learning.
\begin{table}[t]
\centering
\caption{Performance comparison of five emotion categories on the our proposed dataset.}
\label{emotion_5_comparison}
\begin{tabular}{lcccc}
\toprule
\textbf{Emotion} & \textbf{MLP} \cite{palo2015use} & \textbf{GINFormer} \cite{rampasek2022GPS}& \textbf{GIN} \cite{xu2018powerful} &\textbf{SAGE} \cite{hamilton2017inductive} \\
\midrule
Happy & 33.50\% & 46.50\% & 40.00\% & 49.50\%\\
Sad & 40.50\% & 31.50\% & 27.00\% & 27.00\%\\
Fear & 31.50\% & 45.50\% & 29.50\% & 19.00\%\\
Neutral & 27.50\% & 23.50\% & 18.00\% & 43.00\%\\
Angry & 7.00\% & 18.50\% & 36.50\% & 25.50\%\\
\bottomrule
\end{tabular}
\end{table}

The Table \ref{emotion_5_comparison} presents a comparison of performance across different methods on a proposed dataset. The methods compared include MLP, GINformer, GIN, and SAGE, with their performances reported in percentages for five different emotions: Happy, Sad, Fear, Neutral, and Angry. 

SAGE outperforms the other methods in recognizing Happy emotions with a 49.50\% accuracy rate. For Sad emotions, MLP leads with 40.50\%, while for Fear, GINformer shows the highest performance at 45.50\%. In recognising Neutral emotions, SAGE is the best performer with 43.00\% accuracy, and for Angry emotions, GIN achieves the highest accuracy at 36.50\%.

\begin{figure*}
    \centering
    \includegraphics[width=0.9\linewidth]{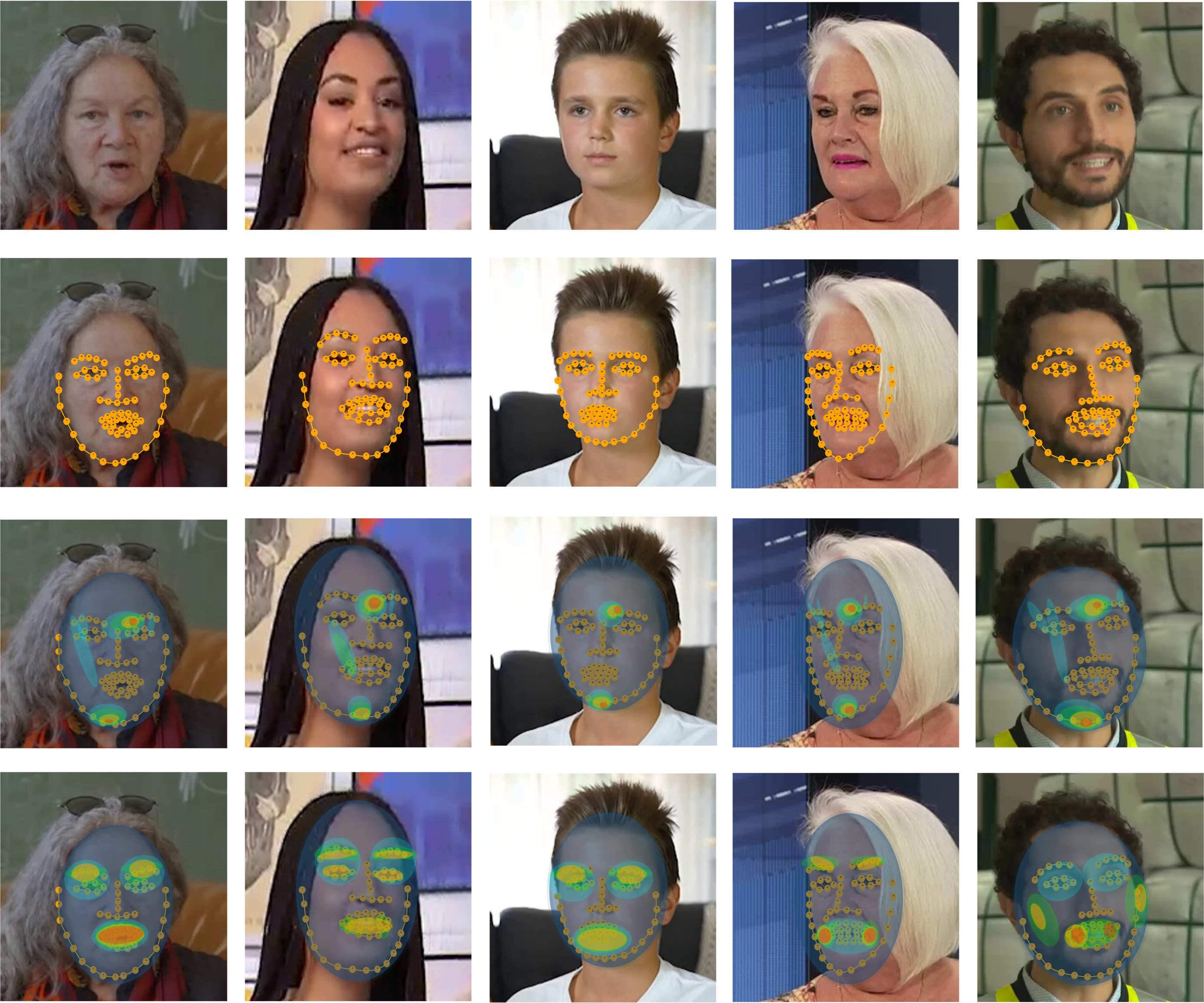}
    \caption{Five emotional landscapes on transformer attention. }
    \label{five_emotion_label}
\end{figure*}

The figure in \ref{five_emotion_label} provides an illustrative comparative study of facial expression analysis using both advanced deep learning algorithms and classical psychological techniques. The first row displays a series of original images portraying a range of emotions—anger, fear, neutrality, sadness, and happiness—extracted from various news broadcasts. The second row showcases the application of facial landmark detection by utilizing a system like MediaPipe\cite{lugaresi2019mediapipe}, which maps out key facial points such as the corners of the mouth or eyes that are essential for analyzing expressions. The third row introduces a layer of complexity with the inclusion of a GINFormer's \cite{rampasek2022GPS} attention mechanism, a state-of-the-art deep learning model that employs transformer architecture to decode the web of facial landmarks and their connections, pinpointing the areas most relevant for identifying each emotion. The fourth row adopts a psychological lens, employing the Facial Action Coding System (FACS) to focus on the activation of Action Units (AUs)—specific facial muscles associated with emotion display. The patterns overlaying the faces in this row indicate the intensity and activity of AUs, offering a more granular perspective on how emotions manifest physically.

The differences of the third and fourth rows illuminates the methodological divergence between computational and psychological analysis: the GINFormer model abstracts emotion recognition into patterns within the facial landmark network, while psychological analysis, through FACS, delves into the intricate details of muscle movements linked with emotional states. This comparative visualization underscores the distinctive, yet potentially complementary, natures of these approaches. Deep learning models like GINFormer excel in identifying and leveraging facial feature patterns to deduce emotions, while psychological methods provide an in-depth, muscle-by-muscle dissection of facial expressions as per the FACS. Such findings highlight the synergy between AI's objective pattern recognition and the qualitative, detailed muscle movement analysis at the heart of psychological research, suggesting a multidisciplinary fusion that could enrich the field of emotion recognition.

%%%%%% 7 emotion categories
% \begin{table}[h]
% \centering
% \caption{Emotion-specific performance comparison between MLP and GANet on the KDEF dataset.}
% \label{tab:emotion_comparison}
% \begin{tabular}{lcccc}
% \toprule
% \textbf{Emotion} & \textbf{MLP Accuracy} & \textbf{GINFormer} & \textbf{GIN} &\textbf{SAGE} \\
% \midrule
% Surprise & 0\% & 0\% & 12 \% & 0\%\\
% Happy & 32\% & 91\% & 39\% & 35\%\\
% Sad & 29\% & 29\% & 16\% & 28\%\\
% Fear & 1\% & 2\% & 11\% & 1\%\\
% Neutral & 58\% & 9\% & 61\% & 67\%\\
% Angry & 0\% & 0\% & 12\% & 0\%\\
% Disgust & 0\% & 0\% & 1\% &0 \%\\
% \bottomrule
% \end{tabular}
% \end{table}

% \begin{table}[h]
% \centering
% \caption{Emotion-specific performance comparison between MLP and GANet on the KDEF dataset.}
% \label{tab:emotion_comparison}
% \begin{tabular}{lcccc}
% \toprule
% \textbf{Emotion} & \textbf{MLP} & \textbf{GINFormer} & \textbf{GIN} &\textbf{SAGE} \\
% \midrule
% Average Accuracy & 31.43\% & 25.71\% & 40.00\% & 28.57\%\\
% \bottomrule
% \end{tabular}
% \end{table}
\section{Conclusion}
In this study, we propose a novel benchmark using realistic news videos, complemented with RGB images and facial landmark coordinates, to enhance emotion recognition research. Our benchmark provides detailed emotion labels and facial landmark data, proving to be a reliable and effective tool in our evaluations. We have demonstrated the practicality and reliability of our dataset, making it a valuable asset for understanding and analyzing emotions. We believe this work will inspire further research in emotion recognition, landmark analysis, and psychological studies. Our contributions aim to support and advance these fields, offering a robust dataset that reflects real-world complexity and variability in human emotions.

\bibliographystyle{IEEEtran} 
\bibliography{2023wow/refswow}    

\end{document}